# Constructing Conditional Plans by a Theorem-Prover

**Jussi Rintanen**                                            RINTANEN@INFORMATIK.UNI-ULM.DE
*Universität Ulm*
*Fakultät für Informatik*
*Albert-Einstein-Allee*
*89069 Ulm, GERMANY*

## Abstract

The research on conditional planning rejects the assumptions that there is no uncertainty or incompleteness of knowledge with respect to the state and changes of the system the plans operate on. Without these assumptions the sequences of operations that achieve the goals depend on the initial state and the outcomes of nondeterministic changes in the system. This setting raises the questions of how to represent the plans and how to perform plan search. The answers are quite different from those in the simpler classical framework. In this paper, we approach conditional planning from a new viewpoint that is motivated by the use of satisfiability algorithms in classical planning. Translating conditional planning to formulae in the propositional logic is not feasible because of inherent computational limitations. Instead, we translate conditional planning to quantified Boolean formulae. We discuss three formalizations of conditional planning as quantified Boolean formulae, and present experimental results obtained with a theorem-prover.

## 1. Introduction

The purpose of automated planning is to construct instructions, a plan, by following which some predefined goals can be achieved. Plans consist of operators that make a set of facts true whenever their preconditions are fulfilled. The most basic – and the most common in earlier research – form of plans is sequence of operators that are executed unconditionally in the specified order. Plans of this form are sufficient only if the world where a plan is carried out is completely predictable and known, and the execution of the plan always starts in the same state.

When not all changes in the world can be predicted or not all facts affecting plan execution are known in advance, the structure of plans has to be more general. If the task is to move object A, that is in room 1 or in room 2, to a trash can, the operations that achieve the goal depend on the initial location of A. There is no single sequence of operations that achieves the goal in both cases. Hence parts of the plan have to be conditional on contingent facts of the world. A plan that achieves the goal says that first go to room 1, if object A is not there go to room 2, pick up the object, find a trash can, and drop the object in it. When following this plan, room 2 is visited only if object A is not in room 1.

Most of the recent work on conditional planning has been carried out in the least-commitment or partial-order planning paradigm, the underlying idea of which has perhaps best been explicated in the planning algorithm SNLP (McAllester & Rosenblitt, 1991). This algorithm starts with an incomplete plan that consists of descriptions of the goal and the initial state. Plans are found by backtracking search. The children of a node in the search tree are generated by extending the incomplete plan. The extensions correspond to fulfilling a subgoal by introducing a new operation or stating that an existing operation is used to fulfill it, and removing potential conflicts between operations by imposing constraints on their ordering.

The conditional planning algorithms CNLP (Peot & Smith, 1992) and Cassandra (Pryor & Collins, 1996) extend partial-order planning algorithms that find non-conditional plans. The nondeterminism and multiple initial states are represented as operators that have several alternative outcomes. Plan search in these algorithms proceeds like in the non-conditional basis algorithm until an operator with more than one outcome is introduced to an incomplete plan. Then a context mechanism that handles conditionality is applied: if there are $n$ alternative outcomes, each current goal is replaced by $n$ new ones each with a different label corresponding to one of the alternative outcomes. The plan is complete when every goal and subgoal – with all existing context labels – is fulfilled.

In this paper we consider conditional planning from a more abstract point of view. Instead of extending existing algorithms that produce non-conditional plans, we view conditional planning as an automated reasoning task, like in the pioneering work on planning by Green (1969), in deductive planning (Rosenschein, 1981), and in recent work on planning by satisfiability algorithms (Kautz & Selman, 1992, 1996). Instead of using a very general framework like the first-order predicate logic or a dynamic logic, we choose a logic that is sufficiently expressive for representing conditional planning but also restricted enough to have potential for efficient implementation. The efficiency requirement together with the recent success of satisfiability algorithms in classical planning (Kautz & Selman, 1996) would suggest that translating problem instances of conditional planning to sets of formulae in the propositional logic and then finding conditional plans by a satisfiability algorithm would be a reasonable way to proceed. However, this turns out not to be the case.

We show that viewing conditional planning as a satisfiability problem is not feasible. Even with the restriction to plans that have a polynomial length the problem of testing the existence of conditional plans almost certainly does not belong to the complexity class NP. Planning by satisfiability consists of constructing a candidate plan (a sequence of operations) and a polynomial-time verification that the operations in the plan achieve the goal when starting from the initial state. Conditional planning, however, involves constructing a plan (*there exists* a plan) such that *for every* combination of contingencies *there exists* an execution that achieves the goal. This alternation of quantifiers $\exists\forall\exists$ takes conditional planning outside NP.

We propose an approach to conditional planning that is based on translation to a computational problem that is a generalization of satisfiability of propositional formulae. This problem is the evaluation of truth-values of quantified Boolean formulae. Quantified Boolean formulae characterize the levels of the polynomial hierarchy (Balcázar, Díaz, & Gabarró, 1995) like propositional formulae characterize the problems in the complexity class NP. For example, the truth of quantified Boolean formulae with the prefix $\exists\forall$ is a complete problem for the complexity class $\Sigma_2^p$. As we will show, determining existence of solutions for conditional planning is $\Pi_2^p$-hard. Because – under standard complexity-theoretic assumptions – there is no polynomial time translation from conditional planning to propositional satisfiability, it is not feasible to solve it by an algorithm that finds satisfying assignments of propositional formulae.

The paper is organized as follows. In Section 3 we discuss the results on the computational complexity of conditional planning, which are the motivation for the approach we have chosen. Section 4 presents different translations of conditional planning to quantified Boolean formulae. The more general framework of conditional planning allows more degrees of freedom in choosing what kind of plans a planner produces. We consider a general formalization in which internal state transitions of a plan are described as finite automata, and less general formalizations with more restricted transition functions. For all formalizations we present translations of problem instances to quantified Boolean formulae. A quantified Boolean formula that illustrates the translations is given in Section 5. We have solved a number of simple problems in conditional planning by using a theorem-prover for QBF we have developed. The theorem-prover is briefly discussed in Section 6 and the experiments in Section 7. Finally, in Section 8 we discuss earlier work that is related to ours.

## 2. Preliminaries

Quantified Boolean formulae are of the form $q_1 x_1 q_2 x_2 \cdots q_n x_n \phi$ where $\phi$ is an unquantified propositional formula and the prefix consists of universal $\forall$ and existential $\exists$ quantifiers $q_1, \ldots, q_n$ and the propositional variables $x_1, \ldots, x_n$ that occur in $\phi$. Define $\phi[\psi/x]$ as the formula that is obtained from $\phi$ by replacing occurrences[1] of the propositional variable $x$ by the formula $\psi$. The truth of quantified Boolean formulae is defined recursively as follows. The truth of a formula that does not contain variables, that is, that consists of the constants true $\top$ and false $\bot$ and connectives, is defined in the obvious way by truth-tables for the connectives. A formula $\exists x \phi$ is true if and only if $\phi[\top/x]$ or $\phi[\bot/x]$ is true. A formula $\forall x \phi$ is true if and only if $\phi[\top/x]$ and $\phi[\bot/x]$ are true. Examples of true quantified Boolean formulae are $\forall x \exists y (x \leftrightarrow y)$ and $\exists x \exists y (x \wedge y)$. The formulae $\exists x \forall y (x \leftrightarrow y)$ and $\forall x \forall y (x \vee y)$ are false. Changing the order of two consecutive variables quantified by the same quantifier does not affect the truth-value of the formula. It is often useful to ignore the ordering of consecutive variables and view each quantifier as quantifying a set of formulae, for example $\exists x_1 x_2 \forall y_1 y_2 \Phi$. The size of a quantified Boolean formula can be defined as the number of occurrences of propositional variables in it.

The interest in quantified Boolean formulae in the theory of computational complexity stems from the fact that like propositional satisfiability characterizes the problems in NP, quantified Boolean formulae with different prefixes characterize different classes in the polynomial hierarchy (Balcázar et al., 1995). The complexity class P consists of decision problems that are solvable in polynomial time by a deterministic Turing machine. NP is the class of decision problems that are solvable in polynomial time by a nondeterministic Turing machine. The class co-NP consists of those problems the complements of which are in NP. In general, the class co-C consists of problems whose complements are in the class C. The polynomial hierarchy PH is an infinite hierarchy of complexity classes $\Sigma_i^p$, $\Pi_i^p$, and $\Delta_i^p$ for all $i \geq 0$ that is defined by using oracle Turing machines in the following way.

$$
\begin{array}{lll}
\Sigma_0^p = P & \Pi_0^p = P & \Delta_0^p = P \\
\Sigma_{i+1}^p = \text{NP}^{\Sigma_i^p} & \Pi_{i+1}^p = \text{co-}\Sigma_{i+1}^p & \Delta_{i+1}^p = P^{\Sigma_i^p}
\end{array}
$$

---

1. We assume that nested quantifiers do not quantify the same variable.

$C_1^{C_2}$ denotes the class of problems that is defined like the class $C_1$ except that oracle Turing machines that use an oracle for a problem in $C_2$ are used instead of Turing machines without an oracle. Oracle Turing machines with an oracle for a problem $B$ are like Turing machines except that they may perform tests for membership in $B$ with constant cost. A problem $L$ is *C-hard* (where C may be NP, co-NP or any of the classes in the polynomial hierarchy) if all problems in the class C are polynomial time *many-one reducible* to it; that is, for all problems $L' \in C$ there is a function $f_{L'}$ that can be computed in polynomial time on the size of its input and $f_{L'}(x) \in L$ if and only if $x \in L'$. We say that the function $f_{L'}$ is a translation from $L'$ to $L$. A problem is *C-complete* if it belongs to the class C and is C-hard.

The truth of quantified Boolean formulae with the prefix $\exists x_1^1 \cdots x_n^1$ is a complete problem for NP= $\Sigma_1^p$, and with prefix $\forall x_1^1 \cdots x_n^1$ the problem is complete for co-NP= $\Pi_1^p$. In general, the truth of formulae with prefix $\forall \exists \forall \cdots$ is $\Pi_i^p$-complete if there are $i-1$ alternations of quantifiers, and $\Sigma_i^p$-complete for prefixes $\exists \forall \exists \cdots$ with $i-1$ alternations.

## 3. Complexity of Conditional Planning

In this section we analyze the complexity of conditional planning. The purpose of the analysis is to justify and motivate the approach to conditional planning we adopt.

A natural approach to conditional planning would be to follow Kautz and Selman (1992, 1996) and to translate problem instances to formulae in the propositional logic, and then find plans by an algorithm that tests the satisfiability of propositional formulae. We show that this approach is not feasible. In addition, our results indicate that quantified Boolean formulae have a sufficient generality for representing conditional planning.

Planning by satisfiability is based on the fact that classical planning, when restricted to plans of polynomial size, belongs to the complexity class NP, and therefore can be translated to any NP-complete problem in polynomial time. An NP-complete problem for which there are several efficient decision procedures is the satisfiability of formulae in the propositional logic. Kautz and Selman (1992) show that translating classical planning to formulae in the propositional logic is straightforward. Solution plans are obtained as satisfying truth-value assignments of the propositional formulae in question.

We show that with the restriction to polynomial size plans conditional planning does not belong to the complexity class NP (assuming that the polynomial hierarchy does not collapse to its first level.) This means that there are no polynomial-time translations from conditional planning to classical planning or to propositional satisfiability. Even the sizes of straightforward translations are exponential. The intuitive reason for conditional planning being outside NP is that problems in NP can be solved by guessing a candidate solution and then verifying in polynomial time that it actually is a solution. For finding a conditional plan one can guess a candidate plan but testing that the plan works under all circumstances cannot in general be performed in polynomial time, as the number of different circumstances may be exponential on the size of the problem instance. These considerations suggest that it is in general not feasible to perform conditional planning by a satisfiability algorithm.

The theorem below shows that the problem of determining whether the goal can be reached from every initial state is one of the most complex problems in the complexity class $\Pi_2^p$, and therefore – very likely – not a member of the complexity class NP. The reachability of the goal from all initial states is equivalent to the existence of conditional plans only for sufficiently expressive notions of conditional plans. For more restricted kinds of conditional plans the existence of separate classical plans for every initial state does not guarantee the existence of a conditional plan covering all initial states. In these cases proofs of $\Pi_2^p$-hardness of plan existence have to be different, of course provided that the problems are indeed $\Pi_2^p$-hard and not easier. Existence of classical plans for all initial states does not necessarily imply the existence of conditional plans for example when it is not possible to combine any two plans that work under different circumstances to a conditional plan that works correctly whenever one of the constituent plans does.

**Definition 1** *A problem instance in conditional planning is a triple $\langle I, O, G \rangle$ where $I$ is a set of literals (characterizing the initial states), members of $O$ are pairs $p \Rightarrow e$ (the operators) where $p$ and $e$ are sets of literals, and $G$ is a literal (the goal).*

*A problem instance has a solution if for every initial state (a propositional model) $M$ such that $M \models I$ there is a sequence $o_1, \ldots, o_n$ of operator applications that transform the initial state $M$ to a state $M'$ such that $M' \models G$. The application of $p \Rightarrow e$ in $M_i$ means that $M_i \models p$ and for the following state $M_{i+1} \models e$ and for all propositional variables $v$ that do not occur in $e$, $M_i \models v$ if and only if $M_{i+1} \models v$.*

**Theorem 2** *The problem of existence of solutions for problem instances in conditional planning is $\Pi_2^p$-hard.*

*Proof:* We show that for any quantified Boolean formula with prefix $\forall \exists$ there is a problem instance in conditional planning such that the formula is true if and only if the problem instance has a solution. For the quantified formula to be true, for all truth-values for the universal variables it has to be possible to assign truth-values to the existential variables so that the unquantified formula evaluates to true. In the planning setting this can be simulated as follows. The universal variables correspond to facts in the initial state that can be both true or false. For each initial state there has to be a sequence of operations that assigns the existential variables truth-values that make the unquantified formula true.

Let $F$ be any quantified Boolean formula $\forall x_1 \cdots x_n \exists y_1 \cdots y_m \Phi$ where $\Phi$ consists of $t$ clauses. We construct a problem instance $P$ and show that $F$ is true if and only if $P$ has a solution. Constructing $P$ takes polynomial time on the length of $F$. This shows that testing the existence of solutions in conditional planning is $\Pi_2^p$-hard. Define $P = \langle I, O, G \rangle$ where $G = sat$,

$$I = \{\neg sat, \neg y_1, \ldots, \neg y_m, \neg c_1, \ldots, \neg c_t, s\}, \text{ and}$$
$$O = \{s \Rightarrow y_1, \ldots, s \Rightarrow y_m, (c_1, \ldots, c_t) \Rightarrow sat\}$$
$$\cup \{l_i \Rightarrow c_j, \neg s | l_1 \vee \cdots \vee l_k \text{ is } j\text{th clause in } \Phi, 1 \leq i \leq k\}.$$

In the initial states the variables $x_i$ may have any value and variables $y_i$ are all false. The operators $s \Rightarrow y_i$ can be executed to make variables $y_i$ true. The truth of clauses $c$ in $\Phi$ can be verified by executing operators $l \Rightarrow c, \neg s$. Finally, the operator $(c_1, \ldots, c_t) \Rightarrow sat$ can be applied to produce the goal if all clauses are true. The variable $s$ is needed to prevent false formulae appear true, like $\exists p(p \wedge \neg p)$ by the plan $(\neg p \Rightarrow c_2), (\Rightarrow p), (p \Rightarrow c_1), (c_1, c_2 \Rightarrow sat)$.

Assume that for all assignments of truth-values to $x_1, \ldots, x_n$ the formula $\exists y_1 \cdots y_m \Phi$ is true. Take any initial state $M$ that satisfies $I$. Now $M$ determines an assignment of truth-values $b_1, \ldots, b_n$ to $x_1, \ldots, x_n$. Let $b'_1, \ldots, b'_m$ be the respective truth-values for $y_1, \ldots, y_m$ as determined by $F$. Let $o_1, \ldots, o_s$ be a sequence consisting of exactly those operators $s \Rightarrow y_i \in O$ such that $b'_i = true$, followed by those operators $l \Rightarrow c_j, \neg s$ such that $l = y_i$ and $b'_i = true$ or $l = \neg y_i$ and $b'_i = false$, and finally $c_1, \ldots, c_t \Rightarrow sat$. Obviously, $o_1, \ldots, o_s$ takes the initial state $M$ to a state $M'$ such that $M' \models sat$. Therefore $P$ has a solution.

Assume $P$ has a solution. Take any assignment $b_1, \ldots, b_n$ of truth-values to $x_1, \ldots, x_n$. Let $M$ be a propositional model such that $M \models I$ and for all $i \in \{1, \ldots, n\}$ $M \models x_i$ iff $b_i = true$. Now there is a sequence $o_1, \ldots, o_s$ of operators taking $M$ to $M'$ such that $M' \models sat$. For all $i \in \{1, \ldots, m\}$ assign $true$ to $y_i$ if $s \Rightarrow y_i$ occurs in $o_1, \ldots, o_s$ before the first $l_i \Rightarrow c_j, \neg s$, and $false$ otherwise. It is easy to show that the assignment to $x_1, \ldots, x_n, y_1, \ldots, y_m$ satisfies $\Phi$. Therefore $F$ is satisfiable. $\square$

Conditional planning is also no harder than problems on the second level of the polynomial hierarchy. For conditional plans of polynomial size it is easy to show that finding a plan is in the complexity class $\Sigma_2^p$. The proof is by constructing a nondeterministic Turing machine that runs in polynomial time and uses an oracle for a problem in NP. The Turing machine first guesses a polynomial length string that represents a candidate plan. The oracle then checks whether it is the case that under some circumstances the goals cannot be reached. The oracle is represented by a nondeterministic Turing machine that guesses truth-values for all contingent facts (in nondeterministic polynomial time), and then executes the plan (in deterministic polynomial time) and accepts if a goal state was not reached. The computation of the oracle is clearly in NP. As our Turing machine runs in nondeterministic polynomial time with an NP oracle, the problem it solves is in $\Sigma_2^p$.

The complexity of classical planning is known in detail. Bylander (1994) shows that classical planning in finite domains is PSPACE-complete. Possibility of tractable planning under syntactic restrictions on the plan operators has been investigated by Bylander and by Bäckström and Nebel (1995). Tractability can be achieved only with very severe syntactic restrictions.

## 4. Encodings of Conditional Planning as Quantified Boolean Formulae

We have devised several translations of conditional planning to quantified Boolean formulae. There are three separate issues in the translation of conditional planning to quantified Boolean formulae. The first, that is no different from translating classical planning to propositional logic, is the encoding of executions of plans. The correspondence between plan executions and plans is not as close as in classical planning, as one plan may have several different executions. The second issue is the representation of plans. Plans are objects that map a state to the operators to be executed for producing a successor state. The third issue, specific to QBF, is the representation of quantification over all initial states and other uncertainties. In Sections 4.3.1 and 4.3.2 we propose two ways of doing this.

Unlike in classical planning where plans simply specify a sequence of operators that are executed consecutively and the state of the environment after each operation is unambiguously known already at the planning time, conditional plans have to be able to behave differently under different circumstances. For example, the environment may be different on different executions of the plan and there may be nondeterministic events affecting plan execution. The different responses required from conditional plans can be handled by defining conditional plans as objects with an internal state that reflects the current and earlier states of the environment in a sufficient extent so that correct operators can be executed. When devising a representation for conditional plans, the decisions to be made concern how the internal state evolves during plan execution, and how the operators to be executed are determined by the internal state.

The idea of conditional plans as objects with an internal state naturally suggests how the notion of classical plans should be extended. The conditional plans discussed in Section 4.2.1 explicitly represent the state of a plan as an automaton that makes transitions based on observations concerning the environment. The state of the automaton at each time point determines which operators are executed. Simpler forms of conditional plans are presented in Sections 4.2.2 and 4.2.3.

The two forms of uncertainty, multiple initial states and nondeterminism, are both important and naturally arise in many applications. However, for simplicity of presentation we postpone the discussion on representing nondeterminism to Section 4.4, and first consider only problem instances with several initial states.

Problem instances $\langle O, B, \Upsilon, \Gamma \rangle$ consist of

1. a finite set $O$ of operators of the form $p \Rightarrow e$ where $p$ and $e$ are finite sets of literals,

2. a set $B$ of *observable facts* that determine how plan execution proceeds,

3. a formula $\Upsilon$ characterizing the initial states, and

4. a formula $\Gamma$ characterizing the goal states.

We assume that the number of atomic facts is finite. Define $prec(p \Rightarrow e) = p$ and $postc(p \Rightarrow e) = e$. Define the size $sizeof(O)$ of a set $O$ of operators as the sum $\sum \{ |prec(o)| + |postc(o)| \mid o \in O \}$. We assume that each operator and fact is assigned a unique integer. The set of integers assigned to operators is denoted by $I_O$, that of facts by $I_F$, and that of observable facts by $I_B$, and $prec(i)$ and $postc(i)$ for $i \in I_O$ have the obvious meaning. We often identify an operator or a fact with its index. Define $N_o = |O|$.

A conditional plan determines for all initial states and combinations of other contingencies an execution that reaches a goal state. This idea is the basis of the representation of conditional planning as quantified Boolean formulae $\exists P \forall C \exists E \Phi$, where $P$ is the set of propositional variables that represent plans, variables in $C$ represent the initial states and other contingencies, and variables in $E$ represent executions of plans. The formula $\Phi$ is a conjunction of formulae that formalize the logical connections between propositions representing plan executions, plans, and initial and goal states.

The propositional variables used in encoding conditional planning are described in Table 1. The representation of classical planning in the framework of Kautz and Selman (1992, 1996) uses the variables $O_{i,t}$ and $P_{i,t}$ only. A propositional variable $(l)_t$ represents the truth of literal $l$ at time point $t$. For a positive literal $l = p$, $(l)_t = P_{i,t}$ where $i$ is the index of $p$, and for a negative literal $l = \neg p$, $(l)_t = \neg P_{i,t}$.

Like in planning by satisfiability (Kautz & Selman, 1992), plans usually cannot be found by performing only one call to a theorem-prover with one formula. This is because the problem encodings depend on the plan size, and there is no obvious upper bound for it. Therefore the theorem-prover is first called with a formula that encodes the smallest interesting plan, and the size is gradually increased until a plan is found. Plan size can be characterized by the length of its executions $t_{max}$ and the number of internal states it may be in. Plan existence corresponds to the truth of the quantified Boolean formula in question, and the plan is represented by the truth-values of propositional variables that represent plan elements.

| variable | description |
|----------|-------------|
| $P_{i,t}$ | The fact $i$ is true at time $t$ (like in satisfiability planning.) |
| $O_{i,t}$ | The operator $i$ is executed at time $t$ (like in satisfiability planning.) |
| $C_{i,j}$ | In state $i$, proposition $j$ (the condition) determines the successor state. |
| $S_i S_j T$ | The successor of state $i$ is $j$ if the condition is *true*. |
| $S_i S_j F$ | The successor of state $i$ is $j$ if the condition is *false*. |
| $S_{i,t}$ | The plan is in state $i$ at time $t$. |
| $E_{i,t}$ | The operator $i$ is enabled at time $t$ (Section 4.2.3) or in state $t$ (Sections 4.2.1 and 4.2.2); that is, it is executed if its preconditions are true. |
| $A_{i,t}$ | The operator $i$ is applicable at $t$; that is, it is enabled, its preconditions are true, and some of its postconditions are false. |

Table 1: Meaning of propositional variables in the encodings

To illustrate the translations, we interleave the presentation of the encodings with examples on encoding a simple blocks world problem. There are two blocks, A and B. The blocks may be on the table or on the top of the other block. To represent this scenario we use the facts *ontableB*, *onBA*, *clearB*, *ontableA*, *clearA*, and *onAB*. The blocks may be moved as specified by the following four operators.

$$0: \quad onAB, clearA \Rightarrow ontableA, \neg onAB, clearB$$
$$1: \quad onBA, clearB \Rightarrow ontableB, \neg onBA, clearA$$
$$2: \quad ontableA, clearB \Rightarrow onAB, \neg clearB, \neg ontableA$$
$$3: \quad ontableB, clearA \Rightarrow onBA, \neg clearA, \neg ontableB$$

## 4.1 Representation of Executions of Plans

Given a sequence of (sets of) operators determined by a plan and an initial state, the execution of the plan involves producing a sequence of successor states, the last of which should be a goal state. To represent plan executions we need formulae that describe the initial states and produce for each state a successor state that corresponds to the application of the operators determined by the plan. In classical planning the plan explicitly gives a set of operators to be executed at each point of time, whereas in conditional planning the set of operators may depend on truth-values of facts or the internal state of the plan. We discuss the formulae that determine the operators to be executed later together with the different formalizations of conditional plans, as the former depend on the latter.

Plan executions are represented like classical planning as a satisfiability problem (Kautz & Selman, 1992). This is because in classical planning plans coincide with their unique executions: both are sequences of (sets of) operators. Plan executions are formalized as formulae that state the preconditions and postconditions of operators (schema 1.1) and frame axioms that say when facts retain their truth-values (schema 1.2).

$$(1.1) \ O_{i,t} \to ((l_1)_t \wedge \cdots \wedge (l_n)_t \wedge (l'_1)_{t+1} \wedge \cdots \wedge (l'_{n'})_{t+1})$$
$$(1.2) \ (l)_t \vee \overline{(l)_{t+1}} \vee O_{n_1,t} \vee \cdots \vee O_{n_m,t} \quad \text{for all } t \in \{0, \ldots, t_{max} - 1\}$$

For operators $i \in I_O$ in schema 1.1, $prec(i) = \{l_1, \ldots, l_n\}$ and $postc(i) = \{l'_1, \ldots, l'_{n'}\}$. The frame axiom says that if literal $l$ is false at $t$ and true at $t + 1$, then one of the operators $n_1, \ldots, n_m$ that make $l$ true is executed at $t$.

If we allow the execution of several operators simultaneously, we need formulae that state that two operators are not executed at the same time if they are dependent; that is, if a propositional variable in the postcondition of one occurs in the precondition of the other. If no parallelism is allowed, we have formulae $\neg(O_{i,t} \wedge O_{j,t})$ for $t \in \{0, \ldots, t_{max} - 1\}$ and for all $\{i, j\} \subseteq I_O$ such that $i \neq j$.

The size of the set of formulae obtained from schemata 1.1 and 1.2 is of the order $(|I_F| + sizeof(O))t_{max}$.

**Example 4.1** In the blocks world example, the formulae describing the preconditions and the postconditions of the operators are the following for $t \in \{0, 1\}$. [2]

$$O_{0,t} \to (onAB_t \wedge clearA_t \wedge ontableA_{t+1} \wedge \neg onAB_{t+1} \wedge clearB_{t+1})$$
$$O_{1,t} \to (onBA_t \wedge clearB_t \wedge ontableB_{t+1} \wedge \neg onBA_{t+1} \wedge clearA_{t+1})$$
$$O_{2,t} \to (ontableA_t \wedge clearB_t \wedge onAB_{t+1} \wedge \neg clearB_{t+1} \wedge \neg ontableA_{t+1})$$
$$O_{3,t} \to (ontableB_t \wedge clearA_t \wedge onBA_{t+1} \wedge \neg clearA_{t+1} \wedge \neg ontableB_{t+1})$$

The frame axioms are as follows for $t \in \{0, 1\}$.

$$
\begin{array}{ll}
\neg onAB_t \vee onAB_{t+1} \vee O_{0,t} & \neg onBA_t \vee onBA_{t+1} \vee O_{1,t} \\
onAB_t \vee \neg onAB_{t+1} \vee O_{2,t} & onBA_t \vee \neg onBA_{t+1} \vee O_{3,t} \\
ontableA_t \vee \neg ontableA_{t+1} \vee O_{0,t} & ontableB_t \vee \neg ontableB_{t+1} \vee O_{1,t} \\
\neg ontableA_t \vee ontableA_{t+1} \vee O_{2,t} & \neg ontableB_t \vee ontableB_{t+1} \vee O_{3,t} \\
clearA_t \vee \neg clearA_{t+1} \vee O_{1,t} & \neg clearA_t \vee clearA_{t+1} \vee O_{3,t} \\
clearB_t \vee \neg clearB_{t+1} \vee O_{0,t} & \neg clearB_t \vee clearB_{t+1} \vee O_{2,t}
\end{array}
$$

The simultaneous application of two operators is not allowed, which is represented by the following formulae for all $\{i, j\} \subseteq I_O = \{0, 1, 2, 3\}$ such that $i \neq j$ and for all $t \in \{0, 1\}$.

$$\neg(O_{i,t} \wedge O_{j,t})$$

$\square$

To represent classical planning as a satisfiability problem, as proposed by Kautz and Selman (1992, 1996), in addition to the above formulae it suffices to give a set of literals that describe an initial state and a formula that describes the goals. Then a satisfiability algorithm can be used for finding a truth-value assignment that satisfies the propositional formulae. The truth-values for propositional variables $O_{i,t}, i \in I_O, t \in \{0, t_{max} - 1\}$ indicate which operators should be applied to reach the goals.

---

2. For clarity, instead of using propositions $P_{i,t}$ where $i$ is the index of a fact, we simply attach the subscript $t$ to the names of the facts, for example $onAB_t$.

## 4.2 Representation of Conditional Plans

Conditional plans are objects that map the current and past observations to the operators to be executed. By the Church-Turing thesis, most general computable notions of such mappings are equivalent to Turing machines. However, it is not necessary to consider mappings from arbitrary observations to sequences of operations. We consider only systems that are represented by finite sets of facts, and hence the plans do not have to be able to respond to arbitrarily complex behavior of the environment.

We consider conditional plans that are finite state; that is, in addition to the information obtained as observations, only a finite amount of information internal to the plan is used in determining which operations to perform and how the internal state evolves. The control flow in this kinds of plans is similar to that of finite automata, or equivalently to that of programs in a simple programming language with iteration or a goto-statement and simple if-then-else conditionals. The finite amount of information, that is the internal state of the plan during execution, can be characterized by a state variable that corresponds to a program counter.

Conditional plans with unrestricted transition functions are very expressive but the number of plans with even a small number of states and observable facts is very high, which makes plan search difficult. As there is, in general, a trade-off between the expressivity of the representation and the difficulty of finding plans, we also consider more restricted forms of conditional plans in Sections 4.2.2 and 4.2.3.

### 4.2.1 Plans with Unrestricted Transition Functions

The first formalization of conditional plans uses finite automata for representing the internal state transitions the plan makes. The successor state of a state is determined by the truth-value of an observable fact associated with the state, which we call *the condition* of the state. The transition functions of the automata may be cyclic in the sense that an automaton may return to a state it has once left.

Each state of a conditional plan has an associated set of operators. We say that for a given state, these operators are *enabled* in it. If an operator is enabled in the current state, it is executed if its preconditions are true.

In domains in which only plan execution may cause changes in the environment, this form of plans is sufficient: whenever a problem instance in conditional planning has a solution, that is, there is plan according to some reasonable notion of conditional plans, it has a solution as the kind of plan discussed in this section. For the simpler notions of plans in Sections 4.2.2 and 4.2.3 this is not the case (see Example 4.3 in Section 4.2.2.)

The number of automata with even a small number of states is fairly high and there is no a priori upper bound on the number of states needed, so parameterizing the encoding with respect to the number $N_s$ of states is necessary. Solutions are first sought for by running a theorem-prover with encodings with a small number of states and points of time, and then gradually increasing the values of these parameters.

The formulae for formalizing conditional plans of this form are given in Figure 1. Define $I_S = \{1, \ldots, N_s\}$. Schemata 2.1 and 2.2 state that for every state there is exactly one condition. Schemata 3.1-3.4 state that for every state there is exactly one successor state for both the true and the false value of the condition. This is needed to ensure that the

Uniqueness of conditions

(2.1) $C_{i,j} \rightarrow \neg C_{i,k}$           for all $i \in I_S$, $\{j,k\} \subseteq I_B$ such that $j \neq k$,

(2.2) $C_{i,n_1} \vee C_{i,n_2} \vee \cdots \vee C_{i,n_m}$    and an enumeration $n_1, \ldots, n_m$ of $I_B$

Uniqueness of successor states in the transition function

(3.1) $S_i S_j T \rightarrow \neg S_i S_k T$    for all $\{i,j,k\} \subseteq I_S$ such that $j \neq k$

(3.2) $S_i S_j F \rightarrow \neg S_i S_k F$

(3.3) $S_i S_1 T \vee S_i S_2 T \vee \cdots \vee S_i S_{N_s} T$    for all $i \in I_S$

(3.4) $S_i S_1 F \vee S_i S_2 F \vee \cdots \vee S_i S_{N_s} F$

Starting state

(4.1) $S_{1,0}$

Uniqueness of current state

(5.1) $S_{i,t} \rightarrow \neg S_{j,t}$   for all $t \in \{0, \ldots, t_{max}\}$, $\{i,j\} \subseteq I_S$ such that $i \neq j$

Transition to a successor state

(6.1) $S_{i,t} \wedge C_{i,k} \wedge P_{k,t} \wedge S_i S_j T \rightarrow S_{j,t+1}$      for all $t \in \{0, \ldots, t_{max}-1\}$,

(6.2) $S_{i,t} \wedge C_{i,k} \wedge \neg P_{k,t} \wedge S_i S_j F \rightarrow S_{j,t+1}$      $\{i,j\} \in I_S, k \in I_B$

Application of operators

(7.1) $(E_{i,j} \wedge S_{j,t} \wedge (l_1)_t \wedge \cdots \wedge (l_n)_t) \rightarrow O_{i,t}$     for all $j \in I_S, i \in I_O$,

(7.2) $O_{i,t} \rightarrow ((E_{i,1} \wedge S_{1,t}) \vee \cdots \vee (E_{i,N_s} \wedge S_{N_s,t}))$    $t \in \{0, \ldots, t_{max}-1\}$ and where $prec(i) = \{l_1, \ldots, l_n\}$

Figure 1: Encoding of conditional plans with unrestricted transition functions

transition functions $I_S \times \{\top, \bot\} \rightarrow I_S$ of plans are well-defined. Formula 4.1 states that the plan execution starts in state 1 at time 0. The choice for state 1 is arbitrary, and obviously does not sacrifice generality. Schema 5.1 states that the plan cannot be in two states at the same time. Schemata 6.1 and 6.2 choose the successor state on the basis of the truth-value of the condition and the transition relation. Schemata 7.1 and 7.2 apply exactly those operators that are enabled in the current state and for which the preconditions are true.

The sizes of the formulae represented by the schemata in Figure 1 are $N_s(|B|^2 - |B|) + N_s|B| + 2(N_s^3 - N_s^2) + 2N_s^2 + 1 + (N_s^2 - N_s)t_{max} + 2N_s^2|B|t_{max} + N_s\,sizeof(O)t_{max} + N_o N_s t_{max}$. Hence the size of the whole set of formulae is of order

$$N_s{}^3 + N_s|B|^2 + N_s{}^2|B|t_{max} + N_s\,sizeof(O)t_{max}.$$

**Example 4.2** For the blocks world example, we produce the encoding for plans with two states. The schemata 2.1-3.4 yield the following formulae for $s \in \{1, 2\}$. We assume that the facts $ontableA$, $clearA$ and $onAB$ are observable.

$$
\begin{array}{lll}
C_{s,ontableA} \to \neg C_{s,clearA} & S_1 S_1 T \to \neg S_1 S_2 T & S_1 S_1 T \lor S_1 S_2 T \\
C_{s,ontableA} \to \neg C_{s,onAB} & S_1 S_2 T \to \neg S_1 S_1 T & S_2 S_1 T \lor S_2 S_2 T \\
C_{s,clearA} \to \neg C_{s,ontableA} & S_2 S_1 T \to \neg S_2 S_2 T & \\
C_{s,clearA} \to \neg C_{s,onAB} & S_2 S_2 T \to \neg S_2 S_1 T & \\
C_{s,onAB} \to \neg C_{s,clearA} & S_1 S_1 F \to \neg S_1 S_2 F & S_1 S_1 F \lor S_1 S_2 F \\
C_{s,onAB} \to \neg C_{s,ontableA} & S_1 S_2 F \to \neg S_1 S_1 F & S_2 S_1 F \lor S_2 S_2 F \\
C_{s,clearA} \lor C_{s,ontableA} \lor C_{s,onAB} & S_2 S_1 F \to \neg S_2 S_2 F & \\
& S_2 S_2 F \to \neg S_2 S_1 F &
\end{array}
$$

Truth-value assignments that satisfy these formulae represent transition functions of conditional plans. The remaining formulae describe how the plan determines which operators are executed. For $t \in \{0, 1\}$ schemata 4.1 and 5.1 yield the following formulae.

$$
\begin{array}{l}
S_{1,0} \\
S_{1,t} \to \neg S_{2,t} \quad S_{2,t} \to \neg S_{1,t}
\end{array}
$$

Schemata 6.1 and 6.2 that describe state transitions yield for $\{s, s'\} \subseteq \{1, 2\}$ and $t \in \{0, 1\}$ the following formulae.

$$
\begin{array}{l}
S_{s,t} \land C_{s,clearA} \land clearA_t \land S_s S_{s'} T \to S_{s',t+1} \\
S_{s,t} \land C_{s,clearA} \land \neg clearA_t \land S_s S_{s'} F \to S_{s',t+1} \\
S_{s,t} \land C_{s,ontableA} \land ontableA_t \land S_s S_{s'} T \to S_{s',t+1} \\
S_{s,t} \land C_{s,ontableA} \land \neg ontableA_t \land S_s S_{s'} F \to S_{s',t+1} \\
S_{s,t} \land C_{s,onAB} \land onAB_t \land S_s S_{s'} T \to S_{s',t+1} \\
S_{s,t} \land C_{s,onAB} \land \neg onAB_t \land S_s S_{s'} F \to S_{s',t+1}
\end{array}
$$

Schemata 7.1 and 7.2 yield for $t \in \{0, 1\}$ and $s \in \{1, 2\}$ the following formulae.

$$
\begin{array}{ll}
(E_{0,s} \land S_{s,t} \land onAB_t \land clearA_t) \to O_{0,t} & O_{0,t} \to ((E_{0,1} \land S_{1,t}) \lor (E_{0,2} \land S_{2,t})) \\
(E_{1,s} \land S_{s,t} \land onBA_t \land clearB_t) \to O_{1,t} & O_{1,t} \to ((E_{1,1} \land S_{1,t}) \lor (E_{1,2} \land S_{2,t})) \\
(E_{2,s} \land S_{s,t} \land ontableA_t \land clearB_t) \to O_{2,t} & O_{2,t} \to ((E_{2,1} \land S_{1,t}) \lor (E_{2,2} \land S_{2,t})) \\
(E_{3,s} \land S_{s,t} \land ontableB_t \land clearA_t) \to O_{3,t} & O_{3,t} \to ((E_{3,1} \land S_{1,t}) \lor (E_{3,2} \land S_{2,t}))
\end{array}
$$

$\square$

In this formalization conditional plans are determined by the valuation of variables $S_i S_j T$, $S_i S_j F$, $C_{i,j}$ and $E_{j,i}$. To make explicit the meaning of plans there are at least two possibilities. Give a formal definition of conditional plans for example as programs in a simple programming language with conditionals and iteration, or give a mechanism for executing the plans implicitly represented by the valuations of the afore-mentioned variables. We choose the latter alternative because it is more straightforward.

So assume we have a valuation $v : P \to \{\top, \bot\}$ that assigns truth-values to the set $P$ of propositional variables that implicitly represent a conditional plan, and the parameter $t_{max}$ that is the length of the plan execution. The procedure in Figure 2 executes the implicitly

```
t := 0;
s := 1;
WHILE t < t_max DO
   BEGIN
      execute simultaneously all operators i such that v(E_{i,s}) = ⊤
         and the preconditions of operator i are true;
      c := i such that v(C_{s,i}) = ⊤;
      s_T := i such that v(S_s S_i T) = ⊤;
      s_F := i such that v(S_s S_i F) = ⊤;
      IF fact c is true THEN s := s_T ELSE s := s_F;
      t := t + 1;
   END
```

Figure 2: Procedure for executing a plan

represented plan. The word "simultaneously" refers to the requirement that the set of operators $i$ with true preconditions and $v(E_{i,s}) = \top$ has to be identified before any of those operators are executed. Note that by the uniqueness schemata 2.1-3.4 the $i$ in $C_{s,i}$, $S_s S_i T$ and $S_s S_i F$ is uniquely determined by $s$.

### 4.2.2 Plans as Sequences of Sets of Iterated Operators

The plans in the previous section are very general, and the number of even relatively small plans can be very high, which makes finding plans and determining the inexistence of plans difficult. Many conditional planning problems have solutions as more restricted and computationally less expensive forms of plans.

The plans discussed in this section have an internal state like the plans of the previous section, and each state is associated with a set of operators enabled in the state, but the state transitions are more restricted. The plan stays in the same state as long as some of the operators enabled in it are applicable; that is, as long as there is an enabled operator the preconditions of which are true and some of the postconditions are false. When there are no such operators, an unconditional transition to a unique successor state is made.

Not all problem instances that have a solution as a plan of the form discussed in Section 4.2.1 have a solution as a plan of the form discussed in this section.

**Example 4.3** Consider the following operators.

$0 : Bob\text{-}has\text{-}1000DM \Rightarrow \neg Bob\text{-}has\text{-}1000DM, Bob\text{-}in\text{-}Kyoto$
$1 : Bob\text{-}has\text{-}1000DM \Rightarrow \neg Bob\text{-}has\text{-}1000DM, Bob\text{-}in\text{-}Paris$
$2 : food\text{-}in\text{-}Kyoto, Bob\text{-}in\text{-}Kyoto, Bob\text{-}has\text{-}5DM \Rightarrow \neg Bob\text{-}has\text{-}5DM, \neg Bob\text{-}hungry$
$3 : food\text{-}in\text{-}Paris, Bob\text{-}in\text{-}Paris, Bob\text{-}has\text{-}5DM \Rightarrow \neg Bob\text{-}has\text{-}5DM, \neg Bob\text{-}hungry$

Initially exactly one of the facts *food-in-Kyoto* and *food-in-Paris* is true, all of the facts *Bob-has-1000DM*, *Bob-has-5DM* and *Bob-hungry* are true, and the rest of the facts are false. The goal is ¬*Bob-hungry*.

With plans of the form discussed in Section 4.2.1 the goal can be achieved as follows. Initially a transition to one of the internal states 2 and 3 is made, depending on which of

**Definition of applicable operators**

(8.1) $A_{i,t} \leftrightarrow$
  $((l_1)_t \wedge \cdots \wedge (l_n)_t \wedge \neg((l'_1)_t \wedge \cdots \wedge (l'_{n'})_t) \wedge ((E_{i,1} \wedge S_{1,t-1}) \vee \cdots \vee (E_{i,N_s} \wedge S_{N_s,t-1})))$
for all $t \in \{0, \ldots, t_{max} - 1\}, i \in I_O$, where $prec(i) = \{l_1, \ldots, l_n\}, postc(i) = \{l'_1, \ldots, l'_{n'}\}$

**Transition to a successor state**

(9.1) $S_{i,t} \wedge \neg A_{1,t+1} \wedge \cdots \wedge \neg A_{N_o,t+1} \rightarrow S_{i+1,t+1}$
(9.2) $S_{i,t} \wedge (A_{1,t+1} \vee \cdots \vee A_{N_o,t+1}) \rightarrow S_{i,t+1}$  for all $i \in I_S$, $t \in \{0, \ldots, t_{max} - 1\}$

**Starting state**

(10.1) $S_{1,0}$

**Uniqueness of current state**

(11.1) $S_{i,t} \rightarrow \neg S_{j,t}$  for all $t \in \{0, \ldots, t_{max}\}, \{i, j\} \subseteq I_S$ such that $i \neq j$

**Application of operators**

(12.1) $(E_{i,s} \wedge S_{s,t} \wedge (l_1)_t \wedge \cdots \wedge (l_n)_t \wedge \neg((l'_1)_t \wedge \cdots \wedge (l'_{n'})_t)) \rightarrow O_{i,t}$
(12.2) $((l'_1)_t \wedge \cdots \wedge (l'_{n'})_t) \rightarrow \neg O_{i,t}$
(12.3) $((\neg E_{i,s_1} \vee \neg S_{s_1,t}) \wedge \cdots \wedge (\neg E_{i,N_s} \vee \neg S_{N_s,t})) \rightarrow \neg O_{i,t}$
for all $t \in \{0, \ldots, t_{max} - 1\}, i \in I_O$ and $s \in I_S$, and where $prec(i) = \{l_1, \ldots, l_n\}$
and $postc(i) = \{l'_1, \ldots, l'_{n'}\}$

Figure 3: Encoding of conditional plans as sequences of iterated operators

*food-in-Kyoto* and *food-in-Paris* is true. In state 2 the operators 0 and 2 are enabled, in state 3 the operators 1 and 3. Obviously, the goal is achieved in all executions of the plan.

For the form of plans discussed in this section there is no solution to this problem. Initially the operators 0 and 1 are applicable. Only one of these operators can be enabled because under the standard notion of dependency they are dependent, and hence their execution in parallel is not well-defined. Assume operator 0 is enabled and operator 1 is not (the other case is symmetric), and therefore operator 0 is applied first. Now if *food-in-Kyoto* is false and *food-in-Paris* is true, the goal cannot be achieved.  □

Formulae that encode plans are given in Figure 3. Auxiliary variables $A_{i,t}$ defined in schema 8.1 express that the preconditions of an enabled operator $i$ are true at $t$ and some of its postconditions are false. If $A_{i,t}$ is false for all operators $i$, then all applicable operators have been applied and a transition to the next state is made next (schema 9.1), and otherwise execution continues in the same state (schema 9.2.) Formula 10.1 states that the plan execution starts in state 1 at time 0, and schema 11.1 states that the plan cannot be simultaneously in two different states. An operator is applied when it is enabled in the current state, its preconditions are true and some of its postconditions are false (schema 12.1.) If these conditions are not fulfilled, the operator is not applied (schemata 12.2 and 13.3.) An upper bound for the size of the formulae in Figure 3 is $sizeof(O)N_s t_{max} + 2N_o N_s t_{max} + 1 + (N_s^2 - N_s)t_{max} + sizeof(O)N_s t_{max} + sizeof(O)t_{max} + N_o N_s t_{max}$ which is of order

$$N_s^2 t_{max} + N_o N_s t_{max} + sizeof(O)N_s t_{max}.$$

**Example 4.4** For the blocks world the encoding is as follows for $t \in \{0, 1\}$. Schema 8.1 yields the following formulae that indicate when operators are applicable.

$$\begin{aligned}
A_{0,t} &\leftrightarrow (onAB_t \wedge clearA_t \wedge \neg(ontableA_t \wedge \neg onAB_t \wedge clearB_t) \\
&\qquad \wedge ((E_{0,1} \wedge S_{1,t-1}) \vee (E_{0,2} \wedge S_{2,t-1}))) \\
A_{1,t} &\leftrightarrow (onBA_t \wedge clearB_t \wedge \neg(ontableB_t \wedge \neg onBA_t \wedge clearA_t) \\
&\qquad \wedge ((E_{1,1} \wedge S_{1,t-1}) \vee (E_{1,2} \wedge S_{2,t-1}))) \\
A_{2,t} &\leftrightarrow (ontableA_t \wedge clearB_t \wedge \neg(onAB_t \wedge \neg clearB_t \wedge \neg ontableA_t) \\
&\qquad \wedge ((E_{2,1} \wedge S_{1,t-1}) \vee (E_{2,2} \wedge S_{2,t-1}))) \\
A_{3,t} &\leftrightarrow (ontableB_t \wedge clearA_t \wedge \neg(onBA_t \wedge \neg clearA_t \wedge \neg ontableB_t) \\
&\qquad \wedge ((E_{3,1} \wedge S_{1,t-1}) \vee (E_{3,2} \wedge S_{2,t-1})))
\end{aligned}$$

If no operator is applicable, a transition to the successor state is made (schema 9.1), and otherwise the state stays the same (schema 9.2.)

$$\begin{aligned}
S_{1,t} \wedge \neg A_{0,t+1} \wedge \neg A_{1,t+1} \wedge \neg A_{2,t+1} \wedge \neg A_{3,t+1} &\to S_{2,t+1} \\
S_{1,t} \wedge (A_{0,t+1} \vee A_{1,t+1} \vee A_{2,t+1} \vee A_{3,t+1}) &\to S_{1,t+1} \\
S_{2,t} \wedge (A_{0,t+1} \vee A_{1,t+1} \vee A_{2,t+1} \vee A_{3,t+1}) &\to S_{2,t+1}
\end{aligned}$$

Schemata 10.1 and 11.1 yield the following formulae.

$$\begin{aligned}
S_{1,0} \\
S_{1,t} \to \neg S_{2,t} \quad S_{2,t} \to \neg S_{1,t}
\end{aligned}$$

Schema 12.1 yields, for all $s \in \{1, 2\}$, the following formulae that describe when operators are applied.

$$\begin{aligned}
(E_{0,s} \wedge S_{s,t} \wedge onAB_t \wedge clearA_t \wedge \neg(ontableA_t \wedge \neg onAB_t \wedge clearB_t)) &\to O_{0,t} \\
(E_{1,s} \wedge S_{s,t} \wedge onBA_t \wedge clearB_t \wedge \neg(ontableB_t \wedge \neg onBA_t \wedge clearA_t)) &\to O_{1,t} \\
(E_{2,s} \wedge S_{s,t} \wedge ontableA_t \wedge clearB_t \wedge \neg(onAB_t \wedge \neg clearB_t \wedge \neg ontableA_t)) &\to O_{2,t} \\
(E_{3,s} \wedge S_{s,t} \wedge ontableB_t \wedge clearA_t \wedge \neg(onBA_t \wedge \neg clearA_t \wedge \neg ontableB_t)) &\to O_{3,t}
\end{aligned}$$

And finally schemata 12.2 and 12.3 say when operators are not applied.

$$\begin{aligned}
(ontableA_t \wedge \neg onAB_t \wedge clearB_t) \to \neg O_{0,t} &\qquad ((\neg E_{0,1} \vee \neg S_{1,t}) \wedge (\neg E_{0,2} \vee \neg S_{2,t})) \to \neg O_{0,t} \\
(ontableB_t \wedge \neg onBA_t \wedge clearA_t) \to \neg O_{1,t} &\qquad ((\neg E_{1,1} \vee \neg S_{1,t}) \wedge (\neg E_{1,2} \vee \neg S_{2,t})) \to \neg O_{1,t} \\
(onAB_t \wedge \neg clearB_t \wedge \neg ontableA_t) \to \neg O_{2,t} &\qquad ((\neg E_{2,1} \vee \neg S_{1,t}) \wedge (\neg E_{2,2} \vee \neg S_{2,t})) \to \neg O_{2,t} \\
(onBA_t \wedge \neg clearA_t \wedge \neg ontableB_t) \to \neg O_{3,t} &\qquad ((\neg E_{3,1} \vee \neg S_{1,t}) \wedge (\neg E_{3,2} \vee \neg S_{2,t})) \to \neg O_{3,t}
\end{aligned}$$

$\square$

The valuation of the propositional variables $E_{i,s}$ determines a conditional plan. The procedure in Figure 4 executes the implicitly represented plan.

```
t := 0;
s := 1;
WHILE t < t_max DO
  BEGIN
      execute simultaneously all operators i such that v(E_{i,s}) = ⊤ and
        the preconditions of operator i are true and
        some of the postconditions are false;
      IF there is no operator i such that v(E_{i,s}) = ⊤
        and preconditions of i are true and some of the postconditions are false
      THEN s := s+1;
      t := t + 1;
  END
```

<div align="center">Figure 4: Procedure for executing a plan</div>

```
t := 0;
WHILE t < t_max DO
  BEGIN
      execute simultaneously all operators i such that v(E_{i,t}) = ⊤
        and the preconditions of operator i are true;
      t := t + 1;
  END
```

<div align="center">Figure 5: Procedure for executing a plan</div>

### 4.2.3 Plans as Sequences of Sets of Operators

A yet simpler form of plans identifies the internal state of a plan with the current time point, or equivalently, at each time point the plan makes an unconditional transition to a successor state. Because the internal state coincides with time, there is no need for separate state variables that represent internal states of plans, and the encoding is particularly simple, consisting of one schema only.

$$(13.1) \quad O_{i,t} \leftrightarrow (E_{i,t} \wedge (l_1)_t \wedge \cdots \wedge (l_n)_t \wedge \neg((l'_1)_t \wedge \cdots \wedge (l'_{n'})_t))$$

for all $i \in I_O$ and $t \in \{0, \ldots, t_{max}-1\}$ where $prec(i) = \{l_1, \ldots, l_n\}$ and $postc(i) = \{l'_1, \ldots, l'_{n'}\}$. The size of the set of formulae from schema 13.1 is of order $sizeof(O)t_{max}$. The procedure in Figure 5 executes the plan represented by the truth-values of the variables $E_{i,t}$.

**Example 4.5** For the blocks world example the formulae are as follows for $t \in \{0, 1\}$.

$$O_{0,t} \leftrightarrow (E_{0,t} \wedge onAB_t \wedge clearA_t \wedge \neg(ontableA_t \wedge \neg onAB_t \wedge clearB_t))$$
$$O_{1,t} \leftrightarrow (E_{1,t} \wedge onBA_t \wedge clearB_t \wedge \neg(ontableB_t \wedge \neg onBA_t \wedge clearA_t))$$
$$O_{2,t} \leftrightarrow (E_{2,t} \wedge ontableA_t \wedge clearB_t \wedge \neg(onAB_t \wedge \neg clearB_t \wedge \neg ontableA_t))$$
$$O_{3,t} \leftrightarrow (E_{3,t} \wedge ontableB_t \wedge clearA_t \wedge \neg(onBA_t \wedge \neg clearA_t \wedge \neg ontableB_t))$$

<div align="right">□</div>

### 4.3 Quantification over Contingencies

A conditional plan determines – for all combinations of truth-values for propositional variables that represent the initial states and other contingencies – an execution that reaches a goal state. This is represented as the sequence of quantifiers $\exists P \forall C \exists E$ where the first quantifies over plans represented by the variables in the set $P$, the second over all contingencies $C$, and the third over executions $E$. By the truth-definition of quantified Boolean formulae, for each plan $P$ all truth-value assignments to variables in $C$ must be possible. Hence the variables in $C$ must be logically independent. In many cases, however, the truth-values of contingent facts are dependent on each other. For example in the blocks world, if $A$ is on $B$, $B$ cannot be on $A$. To keep the variables in $C$ logically independent, we either cannot quantify over the contingent facts directly or we have to classify assignments to $C$ to those that represent allowed combinations of truth-values and to those that do not, and only require the existence of executions leading to a goal state for the former. We discuss both of these alternatives below.

Let $\Upsilon_0$ be the formula for the initial state with all propositions subscripted with 0, and $\Gamma_{t_{max}}$ similarly the formula for the goal subscripted with $t_{max}$. Let $\Phi$ be a formula representing plans and executions as discussed in Sections 4.1 and 4.2.

#### 4.3.1 QUANTIFYING AUXILIARY VARIABLES

In the first alternative we use auxiliary variables $D_1, \ldots, D_n$ as follows. The formula $\Upsilon_0$ is transformed to an equivalent formula $\Upsilon_0^d$ in disjunctive normal form. If there is a proposition $p$ that occurs in $\Upsilon_0^d$ but does not occur in a disjunct $\phi$, $\phi$ has to be replaced by the disjuncts $\phi \wedge p$ and $\phi \wedge \neg p$. As a result, every disjunct has occurrences of exactly the same variables. Let $\phi_1, \ldots, \phi_m$ be the disjuncts of $\Upsilon_0^d$. The number $n$ of auxiliary variables is the smallest integer greater or equal to $\log_2 m$, so that for every disjunct there is a different truth-value assignment to the auxiliary variables. If $m$ is not a power of two, define $\phi_i = \top$ for all $i \in \{m+1, \ldots, 2^n\}$. Let $Q$ be the formula

$$((D_1 \wedge D_2 \wedge \cdots \wedge D_n) \to \phi_1)$$
$$\wedge((\neg D_1 \wedge D_2 \wedge \cdots \wedge D_n) \to \phi_2)$$
$$\wedge((D_1 \wedge \neg D_2 \wedge \cdots \wedge D_n) \to \phi_3)$$
$$\wedge((\neg D_1 \wedge \neg D_2 \wedge \cdots \wedge D_n) \to \phi_4)$$
$$\vdots$$
$$\wedge((\neg D_1 \wedge \neg D_2 \wedge \cdots \wedge \neg D_{n-1} \wedge \neg D_n) \to \phi_{2^n}).$$

The representation of a problem instance in this case is

$$\exists P \forall C \exists R (Q \wedge \Gamma_{t_{max}} \wedge \Phi)$$

where $C$ consists of the variables $D_1, \ldots, D_n$ and the variables for contingencies without occurrences in $\Upsilon_0$, and $R$ is the set of propositional variables not in $P \cup C$.

**Example 4.6** To represent the quantification over the three initial states we need two auxiliary variables $D_1$ and $D_2$ because $2^1 < 3 \leq 2^2$, or equivalently $1 < \log_2 3 \leq 2$. The problem instance is represented by the following formula (for plans from Section 4.2.3.)

$\exists E_{0,0} \ E_{1,0} \ E_{2,0} \ E_{3,0} \ E_{0,1} \ E_{1,1} \ E_{2,1} \ E_{3,1}$
$\forall D_1 \ D_2$
$\exists onAB_0 \ onBA_0 \ clearA_0 \ clearB_0 \ ontableA_0 \ ontableB_0 \ O_{0,0} \ O_{1,0} \ O_{2,0} \ O_{3,0} \ onAB_1 \cdots$
$(((D_1 \wedge D_2) \rightarrow (clearA_0 \wedge clearB_0 \wedge ontableA_0 \wedge ontableB_0 \wedge \neg onAB_0 \wedge \neg onBA_0))$
$\wedge((D_1 \wedge \neg D_2) \rightarrow (clearA_0 \wedge \neg clearB_0 \wedge \neg ontableA_0 \wedge ontableB_0 \wedge onAB_0 \wedge \neg onBA_0))$
$\wedge((\neg D_1 \wedge D_2) \rightarrow (\neg clearA_0 \wedge clearB_0 \wedge ontableA_0 \wedge \neg ontableB_0 \wedge \neg onAB_0 \wedge onBA_0))$
$\wedge((\neg D_1 \wedge \neg D_2) \rightarrow \top)$
$\wedge onAB_2 \wedge \Phi)$

The outermost existential quantifier quantifies the propositional variables describing plans, the universal quantifier quantifies over the initial states, and the innermost existential quantifier quantifies the rest of the variables that represent executions of the plan for particular values of the universally quantified variables. □

### 4.3.2 Quantifying Contingent Facts Directly

The second alternative quantifies over all values of the propositions $C$ that represent the truth-values of facts in the initial states. If the values represent an initial state, there must be an execution of the plan that produces a goal state. The representation of a problem instance in this case is as follows.

$$\exists P \forall C \exists R ((\Upsilon_0 \rightarrow \Gamma_{t_{max}}) \wedge \Phi)$$

Here $R$ consists of the propositional variables not in $P \cup C$. The formula says that there is a plan which produces a goal state when the execution starts in any of the initial states. Executions corresponding to truth-value assignments to $C$ that do not represent initial states $\Upsilon_0$ do not have to reach the goals.

**Example 4.7** The blocks world example in this case is represented as follows.

$\exists E_{0,0} \ E_{1,0} \ E_{2,0} \ E_{3,0} \ E_{0,1} \ E_{1,1} \ E_{2,1} \ E_{3,1}$
$\forall clearA_0 \ clearB_0 \ ontableA_0 \ ontableB_0 \ onAB_0 \ onBA_0$
$\exists O_{0,0} \ O_{1,0} \ O_{2,0} \ O_{3,0} \ onAB_1 \ onBA_1 onAB_1 \ onBA_1 \ clearA_1 \ clearB_1 \ ontableA_1 \cdots$
$((((clearA_0 \wedge clearB_0 \wedge ontableA_0 \wedge ontableB_0 \wedge \neg onAB_0 \wedge \neg onBA_0)$
$\vee(clearA_0 \wedge \neg clearB_0 \wedge \neg ontableA_0 \wedge ontableB_0 \wedge onAB_0 \wedge \neg onBA_0)$
$\vee(\neg clearA_0 \wedge clearB_0 \wedge ontableA_0 \wedge \neg ontableB_0 \wedge \neg onAB_0 \wedge onBA_0)) \rightarrow onAB_2)$
$\wedge \Phi)$

□

## 4.4 Representation of Nondeterminism

The discussion in the previous sections restricted to only one kind of uncertainty, the possibility of several initial states. However, our framework has a generality that is sufficient for representing other kinds of uncertainties, like nondeterministic operators and nondeterminism in the environment. The changes that are needed concern the frame axioms and the formulae that describe the preconditions and postconditions of operators as discussed in Section 4.1. All other formulae remain unaffected.

Consider an operator with preconditions $l_1, \ldots, l_n$ and two alternative effects $e_1 = l'^1_1, \ldots, l'^1_{n'_1}$ and $e_2 = l'^2_1, \ldots, l'^2_{n'_2}$, one of which is chosen nondeterministically. The nondeterminism in the execution of this operator at time point $t$ is represented by a variable $c_t$ that is one of the universally quantified variables. If the operator is executed at $t$ and $c_t$ is true, then the operator has the effect $e_1$, and if $c_t$ is false, then the effect is $e_2$. The generalization to the case with more than two alternative effects is obvious.

The frame axioms and the formulae describing postconditions of operators have to be rewritten to reflect the nondeterminism. The changes make the effects conditional on the variable $c_t$. Let $i \in I_O$, $prec(i) = \{l_1, \ldots, l_n\}$ and $postc(i) = \{l'_1, \ldots, l'_{n'}\}$.

$$(14.1) \ O_{i,t} \to ((l_1)_t \wedge \cdots \wedge (l_n)_t)$$
$$(14.2) \ O_{i,t} \wedge c_t \to ((l'^1_1)_{t+1} \wedge \cdots \wedge (l'^1_{n'_1})_{t+1})$$
$$(14.3) \ O_{i,t} \wedge \neg c_t \to ((l'^2_1)_{t+1} \wedge \cdots \wedge (l'^2_{n'_2})_{t+1})$$

Frame axioms in general are of the form $(l)_t \vee \overline{(l)_{t+1}} \vee x_1 \vee \cdots \vee x_n$. Here formulae $x_i$ represent the possible causes for the change of $l$ from *false* to *true* at $t$. In Section 4.1 the only possible causes were the applications of operators that make $l$ true. When adopting nondeterministic operators and spontaneous change in the environment, the causes for changes become more complicated. Consider the frame axiom for a literal $l$ that occurs in $e_1$. Now one of the disjuncts $x_j$ in the frame axiom is $O_{i,t} \wedge c_t$ where $i$ is the index of the operator. If $l$ occurs in $e_2$, then $O_{i,t} \wedge \neg c_t$ is one of the disjuncts.

Nondeterministic change can be modeled by rules $p \Rightarrow e_1|e_2$ that work like nondeterministic operators except that one of the alternative effects $e_1$ or $e_2$ becomes true always when the precondition $p$ is true. The formulae describing the effects of this kind of rules are obvious: the truth of $p$ implies the truth of $e_1$ at the next point of time when a variable $c$ representing the nondeterminism is true, and the truth of $e_2$ when $c$ is false. The construction of frame axioms for literals in $e_1$ and $e_2$ is modified like in the case of nondeterministic operators. For a literal $l$ in $e_1$, the frame axiom contains a disjunct that is the conjunction of $p$ and $c$, and for literal $l$ in $e_2$ the respective frame axioms has a disjunct that is the conjunction of $p$ and $\neg c$.

## 5. An Example

The complete quantified Boolean formula for the 2-block example developed in the preceding sections is given next. This formula consists of a formalization of plan executions from Section 4.1, formulae formalizing the plans from Section 4.2.3, and formulae for quantification from Section 4.3.1. The preconditions and the postconditions of the operators are described by the following formulae for $t \in \{0, 1\}$.

$$O_{0,t} \to (onAB_t \wedge clearA_t \wedge ontableA_{t+1} \wedge \neg onAB_{t+1} \wedge clearB_{t+1})$$
$$O_{1,t} \to (onBA_t \wedge clearB_t \wedge ontableB_{t+1} \wedge \neg onBA_{t+1} \wedge clearA_{t+1})$$
$$O_{2,t} \to (ontableA_t \wedge clearB_t \wedge onAB_{t+1} \wedge \neg clearB_{t+1} \wedge \neg ontableA_{t+1})$$
$$O_{3,t} \to (ontableB_t \wedge clearA_t \wedge onBA_{t+1} \wedge \neg clearA_{t+1} \wedge \neg ontableB_{t+1})$$

The frame axioms for $t \in \{0, 1\}$ are as follows.

$$\neg onAB_t \vee onAB_{t+1} \vee O_{0,t} \qquad \neg onBA_t \vee onBA_{t+1} \vee O_{1,t}$$
$$onAB_t \vee \neg onAB_{t+1} \vee O_{2,t} \qquad onBA_t \vee \neg onBA_{t+1} \vee O_{3,t}$$
$$ontableA_t \vee \neg ontableA_{t+1} \vee O_{0,t} \qquad ontableB_t \vee \neg ontableB_{t+1} \vee O_{1,t}$$
$$\neg ontableA_t \vee ontableA_{t+1} \vee O_{2,t} \qquad \neg ontableB_t \vee ontableB_{t+1} \vee O_{3,t}$$
$$clearA_t \vee \neg clearA_{t+1} \vee O_{1,t} \qquad \neg clearA_t \vee clearA_{t+1} \vee O_{3,t}$$
$$clearB_t \vee \neg clearB_{t+1} \vee O_{0,t} \qquad \neg clearB_t \vee clearB_{t+1} \vee O_{2,t}$$

The goal is to have the block A on top of block B after the execution of the plan. This is represented by the atomic formula $onAB_2$. The initial states in the problem consists of all the possible arrangements the blocks A and B can be in: A is on B, B is on A, or both are on the table. This problem has a solution in all three forms of plans in Section 4.2, and we use the one from Section 4.2.3. We get the following formulae for $t \in \{0, 1\}$.

$$O_{0,t} \leftrightarrow (E_{0,t} \wedge onAB_t \wedge clearA_t \wedge \neg(ontableA_t \wedge \neg onAB_t \wedge clearB_t))$$
$$O_{1,t} \leftrightarrow (E_{1,t} \wedge onBA_t \wedge clearB_t \wedge \neg(ontableB_t \wedge \neg onBA_t \wedge clearA_t))$$
$$O_{2,t} \leftrightarrow (E_{2,t} \wedge ontableA_t \wedge clearB_t \wedge \neg(onAB_t \wedge \neg clearB_t \wedge \neg ontableA_t))$$
$$O_{3,t} \leftrightarrow (E_{3,t} \wedge ontableB_t \wedge clearA_t \wedge \neg(onBA_t \wedge \neg clearA_t \wedge \neg ontableB_t))$$

To represent the quantification over the initial states we need two auxiliary variables $D_1$ and $D_2$. The initial states are represented by the following formulae.

$$(D_1 \wedge D_2) \rightarrow (clearA_0 \wedge clearB_0 \wedge ontableA_0 \wedge ontableB_0 \wedge \neg onAB_0 \wedge \neg onBA_0)$$
$$(D_1 \wedge \neg D_2) \rightarrow (clearA_0 \wedge \neg clearB_0 \wedge \neg ontableA_0 \wedge ontableB_0 \wedge onAB_0 \wedge \neg onBA_0)$$
$$(\neg D_1 \wedge D_2) \rightarrow (\neg clearA_0 \wedge clearB_0 \wedge ontableA_0 \wedge \neg ontableB_0 \wedge \neg onAB_0 \wedge onBA_0)$$
$$(\neg D_1 \wedge \neg D_2) \rightarrow \top$$

The outermost existential quantifier quantifies the propositional variables describing plans, the universal quantifier quantifies over the initial states, and the innermost existential variable quantifies the rest of the variables that represent executions of the plan for particular values of the universally quantified variables.

$$\exists E_{0,0} \; E_{1,0} \; E_{2,0} \; E_{3,0} \; E_{0,1} \; E_{1,1} \; E_{2,1} \; E_{3,1}$$
$$\forall D_1 \; D_2$$
$$\exists onAB_0 \; onBA_0 \; clearA_0 \; clearB_0 \; ontableA_0 \; ontableB_0 \; O_{0,0} \; O_{1,0} \; O_{2,0} \; O_{3,0} \; onAB_1 \cdots$$

The quantified Boolean formula with the conjunction of the formulae described earlier as the body and the above quantifiers as the prefix, is true if and only if the problem instance in question has a solution with an execution of three points of time. The solution can be found by a theorem-prover for QBF that returns a truth-value assignment to the outermost existentially quantified variables $E_{0,t}, E_{1,t}, E_{2,t}, E_{3,t}$ for $t \in \{0, 1\}$ that represent plans. One solution assigns *true* to $E_{1,0}$ and $E_{2,1}$, and *false* to all other variables $E_{i,t}$. Hence, at time 0 the block B is moved from top of A onto the table if it is on top of A at 0, and at time 1 the block A is moved from table on the top of block B if A is on the table at time 1. This obviously reaches the goal "A is on top of B" starting from all three initial states.

*PROCEDURE* decide($e, \langle M_1, M_2, \ldots, M_n \rangle, C$)
*BEGIN*
  $C := \text{unit}(C)$;
  *IF* $\emptyset \in C$ *THEN RETURN* false;
  *IF* $C = \emptyset$ *THEN RETURN* true;
  remove all variables not in $C$ from $M_1$;
  *IF* $M_1 = \emptyset$ *THEN RETURN* decide(not $e, \langle M_2, \ldots, M_n \rangle, C$);
  x := a member of $M_1$;
  $M_1 := M_1 \backslash \{x\}$;
  *IF* $e$ *THEN*
    *IF* decide($e, \langle M_1, \ldots, M_n \rangle, C \cup \{x\}$)
    *THEN RETURN* true;
  *ELSE*
    *IF* not decide($e, \langle M_1, \ldots, M_n \rangle, C \cup \{x\}$)
    *THEN RETURN* false;
  *RETURN* decide($e, \langle M_1, \ldots, M_n \rangle, C \cup \{\neg x\}$)
*END*

Figure 6: A decision procedure for quantified Boolean formulae

## 6. The Theorem-Prover

We have developed a theorem-prover for quantified Boolean formulae (Rintanen, 1999) as an extension of the Davis-Putnam procedure for the satisfiability of formulae in the propositional logic (Davis, Logemann, & Loveland, 1962). It is straightforward to extend the Davis-Putnam procedure to handle universal quantifiers; one such extension and some improvements are described by Cadoli et al. (1998). First the variables quantified by the outermost quantifier are considered, then the variables by the second quantifier, and so on. Existential variables correspond to or-nodes in the search tree, universal variables correspond to and-nodes. The basic algorithm is given in Figure 6. The first parameter is *true* if the outermost quantifier with active variables is existential and *false* otherwise, the second parameter represents the quantifiers, and the third is the matrix of the formula in clausal form. For the formula $\exists x_1 x_2 \forall y_1 y_2 \exists x_3 ((x_1 \lor y_1) \land (x_2 \lor y_2 \lor x_3))$ the algorithm is called with arguments ($true, \langle \{x_1, x_2\}, \{y_1, y_2\}, \{x_3\} \rangle, \{x_1 \lor y_1, x_2 \lor y_2 \lor x_3\}$). The subprocedure *unit* performs unit resolution and unit subsumption. The basic algorithm with the standard efficient implementation techniques for the Davis-Putnam procedure is able to solve only simple planning problems, and we have developed new techniques for speeding up the algorithm. Our theorem-prover with the new techniques disabled cannot solve any of the benchmarks in Table 3 in less than 4 hours (14400 seconds.)

The techniques for improving the obvious algorithm are based on failed literal detection (Freeman, 1995; Li & Anbulagan, 1997), and for formulae $\exists X \forall Y \Phi$ on performing computation with the universal variables $Y$ before all variables in $X$ have been assigned a truth-value. First, before performing exhaustive search on all possible truth-value assignments to variables in $X$, assigning any truth-values to some of the variables in $Y$ and then performing unit resolution often yields truth-values to some of the variables in $X$. The truth-values obtained are ones that must be assigned to those variables. Second, at any node of the search tree, if assigning true to $p \in X$ and any truth-values to some of the variables in $Y$ and then performing unit resolution yields a contradiction, then $p$ has to be assigned false. Third, detecting failed literals by unit resolution can also be performed on variables that cannot be chosen as the next branching variable, that is variables quantified by some other than the current outermost quantifier.

An important research topic on reasoning with QBF is how for a true formula with $n$ universally quantified variables, going through all of the $2^n$ truth-value assignments can be avoided. Currently the only technique we employ is partitioning clause sets – at each node of the search tree – so that no two sets have variables in common. For some planning problems, the QBF is split to several clause sets with a low number of universal variables occurring in each, and this way the number of assignments that have to be considered is much less than $2^n$.

## 7. Experiments

We have written a program that translates conditional planning problems to QBF, and performed computational experiments in which conditional plans have been found by the theorem-prover discussed in Section 6.

All the translations are automatically generated from the set of operators and the formulae that describe the initial and goal states. This program includes the automatic generation of invariants for the problem instance. The use of invariants (Gerevini & Schubert, 1998; Rintanen, 1998) and related techniques for pruning search spaces is one of the distinguishing features of recent classical planners. By invariants we mean formulae that are true in all reachable states of a problem instance. Invariants are determined by the operators and the initial states, and they help plan search also in conditional planning.

We have run two series of benchmarks. The first benchmark demonstrates planning under $n$ independent sources of uncertainty that corresponds to $2^n$ problem instances without uncertainty. The purpose of this benchmark is to demonstrate that it can be more natural and much easier to solve the whole conditional planning problem once, instead of solving the corresponding classical problem instances separately. The second benchmark is the blocks world with several initial states. We let our theorem-prover find a conditional plan that reaches a given goal state starting from every possible state. This benchmark is parameterized by the number of blocks, and we can solve the problem with 4 blocks in a reasonable time. For 5, 6 and 7 blocks the problem has respectively 501, 4051 and 37633 initial states, and our translator has difficulties in translating a QBF representation of these states to clausal form.

The first scenario consists of a sequence of rooms with a pair of doors connecting consecutive rooms. Exactly one door of each pair is open, and before executing the plan it is not known which. The goal is to go from the first room to the last. We ran this benchmark with the encoding from Section 4.2.3. In this case there is only one parameter that is increased during plan search, the length of the plan. Hence the theorem-prover is first called with a formula that encodes plans/executions of length 1, then with a formula for length 2, and so on. From the first formula that is found to be true a plan that reaches the goal can be extracted. In Table 2 we give statistics on the evaluation of formulae with the encoding in Section 4.2.3. All the formulae are true and correspond to the shortest plan lengths for which solutions exists. In all cases, the runtimes for evaluating the false formulae that correspond to shorter plan lengths are smaller. The first column in the table gives the number of rooms, the second the number of initial states, the third the number of clauses in the quantified Boolean formula, the fourth the number of propositional variables in that formula, the fifth the runtime, and the sixth the number of non-leaf nodes in the search tree. All the runs were on a Sun Ultra II workstation with a 296 MHz processor. The runtimes with the encoding from Section 4.2.2 are comparable.

| rooms | istates | clauses | vars | runtime | nodes |
|-------|---------|---------|------|---------|-------|
| 13 | 4096 | 4582 | 925 | 0.2 s | 12 |
| 14 | 8192 | 5458 | 1080 | 0.2 s | 13 |
| 15 | 16384 | 6424 | 1247 | 0.3 s | 14 |
| 16 | 32768 | 7483 | 1426 | 0.4 s | 15 |
| 17 | 65536 | 8638 | 1617 | 0.4 s | 16 |
| 18 | 131072 | 9892 | 1820 | 0.6 s | 17 |
| 19 | 262144 | 11248 | 2035 | 0.8 s | 18 |
| 20 | 524288 | 12709 | 2262 | 0.9 s | 19 |
| 21 | 1048576 | 14278 | 2501 | 1.1 s | 20 |
| 22 | 2097152 | 15958 | 2752 | 1.4 s | 21 |
| 23 | 4194304 | 17752 | 3015 | 1.6 s | 22 |
| 24 | 8688608 | 19663 | 3290 | 1.9 s | 23 |

Table 2: Runtimes for the Rooms problem

Table 3 contains statistics on the solution of some blocks world problems. Sample solution plans are given in Appendix A. The goal in these problems is to take the blocks to a certain configuration (a stack containing all the blocks) from all initial configurations. For each of the six problem instances, with three or four blocks and one of the three problem encodings presented earlier, we have generated formulae for several parameter values that characterize the sizes of the solutions. The statistics given in Table 3 are for the true formulae that represent a solution (last in each set of formulae), and the false formulae that correspond to the highest parameter values for which there are no solution plans.

For the runs of the theorem-prover for each formula we give the following information. The first column gives the section in which the encoding is described, the second column the number of blocks, the third the number of initial states, the fourth the length of longest execution needed, the fifth the number of states in the plan (if separate from the number of time points), the sixth the number of clauses in the formula, the seventh the number of propositional variables, the eighth the number of seconds it takes for our theorem-prover to evaluate the formula, the ninth the number of non-leaf nodes in the search tree, and the tenth the truth-value of the formula. With the encoding from Section 4.2.1 the facts *conXY* defined as *onXY ∧ clearX* are observable. Other facts could be used as well, and the choice of facts may affect both the existence of plans and the size of plans.

The runtimes in Table 3 roughly confirm the idea that the plan representations in Sections 4.2.1, 4.2.2 and 4.2.3 are decreasingly computationally demanding. The slightly lower runtime and noticeably lower number of nodes in the search tree for the 4 block example with the plan encoding from Section 4.2.2 than with the simpler encoding from Section 4.2.3 may therefore be surprising. The difference would seem to be due to the fact that the low number of internal states (three) in the plans from Section 4.2.2 forces the plan to stay in some of the states for several points of time. For plans from Section 4.2.3 there is no such restriction and the sets of operators enabled at each point of time may be different. The constraints on solution plans are therefore not as tight for the simpler representation, and more search is needed for finding a plan.

| encoding | blocks | istates | time | states | clauses | vars | runtime | nodes | value |
|----------|--------|---------|------|--------|---------|------|---------|-------|-------|
| S4.2.1 | 3 | 13 | 4 | 4 | 2928 | 288 | 0.1 s | 0 | false |
| S4.2.1 | 3 | 13 | 5 | 3 | 2892 | 286 | 354.4 s | 391 | false |
| S4.2.1 | 3 | 13 | 5 | 4 | 3852 | 328 | 69.7 s | 87 | true |
| S4.2.2 | 3 | 13 | 4 | 3 | 2533 | 247 | 0.1 s | 0 | false |
| S4.2.2 | 3 | 13 | 5 | 2 | 2707 | 282 | 0.6 s | 0 | false |
| S4.2.2 | 3 | 13 | 5 | 3 | 3402 | 304 | 3.4 s | 29 | true |
| S4.2.3 | 3 | 13 | 4 | - | 1433 | 202 | 0.0 s | 0 | false |
| S4.2.3 | 3 | 13 | 5 | - | 1835 | 256 | 1.7 s | 35 | true |
| S4.2.1 | 4 | 73 | 6 | 4 | 15872 | 779 | 51.0 s | 0 | false |
| S4.2.1 | 4 | 73 | 7 | 3 | 15219 | 783 | > 15 h | ? | ? |
| S4.2.1 | 4 | 73 | 7 | 4 | 18768 | 863 | > 15 h | ? | ? |
| S4.2.2 | 4 | 73 | 6 | 3 | 15061 | 838 | 41.1 s | 0 | false |
| S4.2.2 | 4 | 73 | 7 | 2 | 15047 | 915 | 85.9 s | 0 | false |
| S4.2.2 | 4 | 73 | 7 | 3 | 22959 | 1023 | 231.3 s | 190 | true |
| S4.2.3 | 4 | 73 | 6 | - | 9661 | 727 | 11.2 s | 0 | false |
| S4.2.3 | 4 | 73 | 7 | - | 11303 | 855 | 239.0 s | 809 | true |

Table 3: Runtimes for the Blocks World

With these blocks world problem instances the solution of the separate problems is very easy for the best classical planners. What makes these problems difficult is that the plans represent all possible executions, and the constraints on the plans are not as tight as in the benchmarks in Table 2 or in the separate classical planning problems. However, when considering that the number of elements in the resulting plans is relatively high (fifteen or more for the bigger problems) and the notion of plans is much more complicated than in classical planning, the runtimes are not disappointing.

Some of the observations about plan search in our approach are interesting. Even though there are three quantifiers, our theorem-prover does not perform search on variables quantified by the third one that represent plan executions. This is because the plan, represented by the outermost variables, together with the universally quantified variables for the contingencies, uniquely determine the execution that is found without search. This is nicely on par with the fact that conditional planning is on the second level of the polynomial hierarchy, not on the third as the prefix $\exists\forall\exists$ in the encodings might suggest.

As shown in Table 2, the runtimes for plan generation can be much less than linear to the number of initial states. None of the early conditional planning algorithms is able to exhibit similar behavior; that is, they produce plans of exponential length and therefore consume exponential time even on simple problems like these. The favorable runtimes are due to our theorem-prover implementation. For example for the problems in Table 2, naïve extensions of the Davis-Putnam procedure to QBF consider all of the $2^n$ truth-value assignments to the universally quantified variables just to verify that a plan that has been found actually reaches the goal in all cases. Further developments in theorem-proving techniques for QBF and propositional satisfiability are likely to improve these runtimes further.

## 8. Related Work

Both Peot and Smith (1992) and Pryor and Collins (1996) present algorithms for conditional planning that are based on the least-commitment or partial-order planning paradigm. Both algorithms work like corresponding classical planning algorithms until a subgoal is fulfilled by the application of an operator that does not have a unique outcome, that is, the operator is nondeterministic. At that point the development of the conditional plan is split to a number of separate subproblems that are solved separately, each corresponding to one of the outcomes of the nondeterministic operator. The problem with this approach is that the sizes of conditional plans are exponential on the number of uncertainties, and as generating a solution takes at best linear time on the size of the solution (usually exponential), this kind of algorithms inherently consume a lot of computational resources. Furthermore, often the improvement over the trivial conditional planning algorithm that simply reduces the problem to a number of classical planning problems that are solved separately, is small. They fare better than the trivial algorithm whenever some of the contingencies are irrelevant in reaching the goal, or if the separate plans have parts in common.

Cimatti et al. (1998) propose an algorithm for conditional planning that enumerates the state space. Starting from the goal states, the sets of states from which a goal state is reachable with $n \geq 0$ steps or less are computed. When for some $n$ the set includes all initial states, a plan has been found. During the enumeration, each state is associated with an operation that is along a shortest path to a goal state. Now the goals can be reached from any of the initial states by repeatedly applying the operator associated with the current state. As the number of state-action pairs in these plans is as high as the number of states, problem instances with big state spaces consume more memory than is likely to be available. To alleviate this problem Cimatti et al. propose the use of binary decision diagrams (Bryant, 1992) for encoding the state-action tables. BDDs are in general not capable of representing exponential size data structures in polynomial space.

Smith and Weld (1998) extend Graphplan (Blum & Furst, 1997) to handle uncertainty and several initial states. The plans produced by their planner are sequences of operators like in classical planning, but as the effects of operators may be conditional on some facts, the plans may achieve the goals even when starting the plan execution in different states or when there is nondeterminism. Smith and Weld call this conformant planning. Their planning algorithm explicitly represents information on all executions of a plan. This may be possible when the number of initial states is small, up to a couple of dozen or a hundred on small problem instances, but for more complex problems it is not feasible because of high memory consumption. Representing conformant planning in our framework is easy.

Our work and satisfiability planning by Kautz and Selman (1992, 1996) are closely related. A major difference is that we can directly address a much wider range of planning problems with nondeterministic change and several initial states. Because of the added generality, the problems we can solve do not in general belong to the complexity class NP. If all sources of uncertainty are eliminated, our translations contain only existential quantifiers, and these quantified Boolean formulae are true exactly when the same formulae without the quantifiers are satisfiable. Our translations in these cases still contain the more complex representation of conditional plans, but otherwise the resulting sets of formulae are similar and plans can be found by a satisfiability algorithm.

A fundamental difference between the satisfiability algorithms of Kautz and Selman and our theorem-prover is that the former, GSAT and WALKSAT, are based on local search. These algorithms repeatedly try to guess truth-value assignments that satisfy the set of clauses. At no point of time is there a guarantee that all assignments have been considered, and therefore these algorithms are not capable of determining the unsatisfiability of a set of clauses. In general, because local search algorithms do not exhaustively go through the search space, they cannot determine with certainty that an object with a certain property does not exist. A local search algorithm can find a conditional plan but it cannot determine its correctness. Without systematically considering all combinations of contingencies, only counterexamples that show a plan incorrect can be found. Therefore at least those parts of algorithms for conditional planning that verify that a plan is correct have to be systematic.

## 9. Conclusions and Future Work

We propose a new approach to conditional planning that is based on representing problem instances as quantified Boolean formulae and using an automated theorem-prover for finding plans. This approach is both theoretically and practically well motivated. As a practical motivation we see the recent success of satisfiability algorithms (Kautz & Selman, 1996) in classical planning. The problem of determining truth-values of quantified Boolean formulae is a generalization of the problem of satisfiability of propositional formulae. As a theoretical justification we give complexity results that demonstrate that it is in general not feasible to use satisfiability algorithms in conditional planning.

This work differs from earlier work on conditional planning in several respects. Unlike the planning algorithms CNLP and Cassandra (Peot & Smith, 1992; Pryor & Collins, 1996), we do not reduce conditional planning to the simpler case of planning without uncertainties. As shown by our theoretical analysis, this reduction would be ill-motivated, as it most likely cannot be done in polynomial time. In some cases when finding a plan is easy, this reduction makes it very costly. Cimatti et al. (1998) give an algorithm that enumerates the state space of a conditional planning problem. The plans constructed by their algorithm explicitly associate an operation with every state, and this inherently leads to big plans. It is not clear whether the BDD techniques they propose make this feasible for complex problems.

We have developed a prototype implementation of a theorem-prover for QBF and experimented with producing conditional plans with it. The results are preliminary, but give a justification to our approach: simple benchmark problems that would be far too difficult for some of the earlier conditional planners are quickly solved in time that is sublinear on the number of initial states in the problem instances. We believe that expressing conditional planning as a theorem-proving task makes it easier to identify general techniques that benefit the construction of conditional plans, and also to identify techniques that cannot be conveniently embedded in particular theorem-proving frameworks.

```
state 1:
ENABLE clearA onAC  => -onAC ontableA clearC
ENABLE onBA clearB  => -onBA ontableB clearA
ENABLE onBC clearB  => -onBC ontableB clearC
ENABLE onCA clearC  => -onCA ontableC clearA

state 2:
ENABLE onBA clearC clearB  => -onBA -clearC onBC clearA
ENABLE clearA onAC  => -onAC ontableA clearC
ENABLE ontableB clearC clearB  => -ontableB -clearC onBC

state 3:
ENABLE onAB clearA  => -onAB ontableA clearB
ENABLE clearA onAC  => -onAC ontableA clearC
ENABLE onCA clearC  => -onCA ontableC clearA
ENABLE onCB clearC  => -onCB ontableC clearB

state 4:
ENABLE clearA onAC  => -onAC ontableA clearC
ENABLE onCB clearC  => -onCB ontableC clearB
ENABLE ontableA clearA clearB  => -ontableA -clearB onAB
ENABLE ontableB clearC clearB  => -ontableB -clearC onBC
```

Figure 7: Enabled operators for each state

**Acknowledgements**

This research was funded by the Deutsche Forschungsgemeinschaft SFB 527. The author also wishes to thank the Finnish Cultural Foundation and the Finnish Academy of Science and Letters for additional financial support. Holger Pfeifer kindly helped with proof-reading.

**Appendix A: Sample Plans**

Sample plans found by our theorem-prover are given below. In the first problem with three blocks, the goal is to have A on B and B on C. The initial states are all the possible configurations of the three blocks. The plan encoding is the one described in Section 4.2.1 and it has four states. In each state a number of operators are enabled, as described in Figure 7. The execution of the plan starts from state 1, and the goal is reached at time 4. At every point of time a transition to a successor state is made on the basis of truth-values of facts $conXY$ that are defined by $conXY \leftrightarrow (onXY \wedge clearX)$. The transition function is given in Table 4. In all executions of the plan, transitions through the same sequence 1,3,2,4 of states are made.

The second problem instance has four blocks. We give a plan obtained with the encoding in Section 4.2.2. The goal is to have A on B, B on C, and C on D. The initial states are

| transition | when |
|---|---|
| $1 \Rightarrow 3$ | $conBC$ |
| $1 \Rightarrow 3$ | $\neg conBC$ |
| $2 \Rightarrow 2$ | $conAC$ |
| $2 \Rightarrow 4$ | $\neg conAC$ |
| $3 \Rightarrow 2$ | $conAC$ |
| $3 \Rightarrow 2$ | $\neg conAC$ |
| $4 \Rightarrow 2$ | $conAC$ |
| $4 \Rightarrow 1$ | $\neg conAC$ |

Table 4: Transition function

```
state 1:
ENABLE onAB clearA  => -onAB ontableA clearB
ENABLE onAC clearA  => -onAC ontableA clearC
ENABLE onAD clearA  => -onAD ontableA clearD
ENABLE onBA clearB  => -onBA ontableB clearA
ENABLE onBC clearB  => -onBC ontableB clearC
ENABLE onBD clearB  => -onBD ontableB clearD
ENABLE onCA clearC  => -onCA ontableC clearA
ENABLE onCB clearC  => -onCB ontableC clearB
ENABLE onCD clearC  => -onCD ontableC clearD
ENABLE onDA clearD  => -onDA ontableD clearA
ENABLE onDB clearD  => -onDB ontableD clearB
ENABLE onDC clearD  => -onDC ontableD clearC

state 2:
ENABLE onBA clearC clearB  => -onBA -clearC onBC clearA
ENABLE ontableB clearA clearB  => -ontableB -clearA onBA
ENABLE ontableC clearD clearC  => -ontableC -clearD onCD

state 3:
ENABLE ontableA clearA clearB  => -ontableA -clearB onAB
```

Figure 8: Enabled operators for each state

all the possible configurations of the four blocks. The plan has three states. The enabled operators of each state $s$ are applied repeatedly until no more operators are applicable, and then a transition to the successor state $s + 1$ is made. The enabled operators for each state are given in Figure 8. When exiting state 1, all blocks are on the table. In state 2, first B is moved on top of A and C is moved on top of D, and then B is moved from A on top of C. Finally in state 3, A is moved on top of B.